\documentclass{article}

\usepackage{cite}

\usepackage[pdftex]{graphicx}
\DeclareGraphicsExtensions{.pdf,.jpeg,.png}

\usepackage[cmex10]{amsmath}
\usepackage{amsfonts}

\usepackage[caption=false,font=footnotesize]{subfig}

\usepackage{authblk}
\usepackage{enumerate}
\usepackage{color}
\definecolor{red}{rgb}{1, 0, 0}

\newcommand\fullfigwidth{1}
\newcommand\colfigwidth{1}
\allowdisplaybreaks

\DeclareMathOperator*{\argmin}{argmin}
\DeclareMathOperator*{\argmax}{argmax}

\begin{document}

\title{On the Empirical Effect of Gaussian Noise \\ in Under-sampled MRI Reconstruction}
\author{Patrick Virtue and Michael Lustig}
\affil{Department of Electrical Engineering and Computer Sciences, University of California, Berkeley, CA, 94720 USA.}
\date{}

\maketitle

\begin{abstract}

In Fourier-based medical imaging, sampling below the Nyquist rate results in an underdetermined system, in which linear reconstructions will exhibit artifacts. Another consequence of under-sampling is lower signal to noise ratio (SNR) due to fewer acquired measurements. Even if an oracle provided the information to perfectly disambiguate the underdetermined system, the reconstructed image could still have lower image quality than a corresponding fully sampled acquisition because of the reduced measurement time. The effects of lower SNR and the underdetermined system are coupled during reconstruction, making it difficult to isolate the impact of lower SNR on image quality. To this end, we present an image quality prediction process that reconstructs fully sampled, fully determined data with noise added to simulate the loss of SNR induced by a given under-sampling pattern. The resulting prediction image empirically shows the effect of noise in under-sampled image reconstruction without any effect from an underdetermined system.

We discuss how our image quality prediction process can simulate the distribution of noise for a given under-sampling pattern, including variable density sampling that produces colored noise in the measurement data. An interesting consequence of our prediction model is that we can show that recovery from underdetermined non-uniform sampling is equivalent to 
a weighted least squares optimization that accounts for heterogeneous noise levels across measurements.

Through a series of experiments with synthetic and \emph{in vivo} datasets, we demonstrate the efficacy of the image quality prediction process and show that it provides a better estimation of reconstruction image quality than the corresponding fully-sampled reference image.
\end{abstract}

\section{Introduction}

Under-sampling in Fourier-based medical imaging provides a variety of clinical benefits including shorter exam times, reduced motion artifacts, and the ability to capture fast moving dynamics, such as cardiac motion. Under-sampling reduces acquisition time by collecting fewer measurements in the frequency domain than required by the Nyquist rate. However, under-sampling causes two specific challenges for the reconstruction system, namely, an \textbf{underdetermined system}\footnote{In the context of this paper, we specify fully determined and underdetermined as follows: for a fixed Cartesian $k$-space (frequency space) grid with predefined field of view and spatial resolution parameters, \textbf{fully determined} means having at least one measured sample for each $k$-space grid location, and \textbf{underdetermined} means at least one $k$-space location has zero samples, in which case we have more unknowns (image pixels) than equations (one per acquired $k$-space location).}
of linear equations and \textbf{lower SNR} (signal-to-noise ratio) due to reduced measurement time. When reconstruction algorithms are able to overcome these challenges, under-sampling can benefit a variety of Fourier-based imaging modalities, including MRI with parallel imaging or compressed sensing \cite{SMASH}\cite{Sparse_MRI}, computed tomography (CT) with reduced or gated acquisition views \cite{CT_CS_Song}\cite{CT_CS_Sidky}, and positron emission tomography (PET) with multiplexed or missing detectors \cite{PET_CS_Multiplex}\cite{PET_CS_PartialRing}.
While the tools and analysis discussed in this paper apply generally to Fourier-based medical imaging with Gaussian noise, we will direct our numerical modeling, examples, and experiments to the application of compressed sensing MRI.

When designing an under-sampled reconstruction system, the primary concern is often focused on compensating for the underdetermined system caused by sub-Nyquist sampling, for example choosing a sparse representation for compressed sensing.
However, we should not overlook the fact that collecting fewer measurements in practice leads to overall lower SNR in the acquired data.
If the measurements are too noisy, the low SNR will lead to poor reconstructed image quality even if the reconstruction system were fully determined.
On the other hand, with high SNR measurements, the resulting image quality will be limited by how well the reconstruction can constrain the underdetermined system. The effects of the underdetermined system and the lower SNR are coupled during the reconstruction process, making it difficult to analyze one without the other. It is important, however, to analyze how both issues impact the reconstruction system in order to determine the empirical limits of under-sampling and gain insight on how to improve under-sampled acquisition and reconstruction when targeting specific applications.

Compressed sensing theory has provided us with extensive analysis on the bounds for the successful signal recovery from under-sampled data. Cand{\`e}s \cite{CS_RIP} describes a bound on the mean squared error (MSE) of the recovered signal limited by the under-sampling rate and the sparsity of the data. He also shows that this bound scales linearly with the variance of the noise in the measured data. Donoho, et al \cite{CS_PhaseTransition_Noise} extend this analysis by including a bound on the MSE for worst case support of a sparse signal. They demonstrate that the MSE is bounded when the under-sampling rate and sparsity level are below a specific phase transition curve and the MSE is unbounded above this curve, regardless of noise level. Unfortunately, MSE is a fairly limited measure of a successful reconstruction. Wainwright \cite{CS_Noise_SupportRecovery} improves upon the MSE definition of success by studying the under-sampling rates and sparsity levels for which there is a high probability of successfully recovering the support of the sparse signal. However, in practice, the desired signal is often not explicitly sparse and there may be significant information in small values outside of a sparse support. Xu and Hassibi \cite{CS_ApproxSparse} account for this by studying the robustness and bounds of compressed sensing applied to approximately sparse signals.

\begin{figure}[!t]
	\begin{center}
		\includegraphics[width=\colfigwidth\columnwidth]{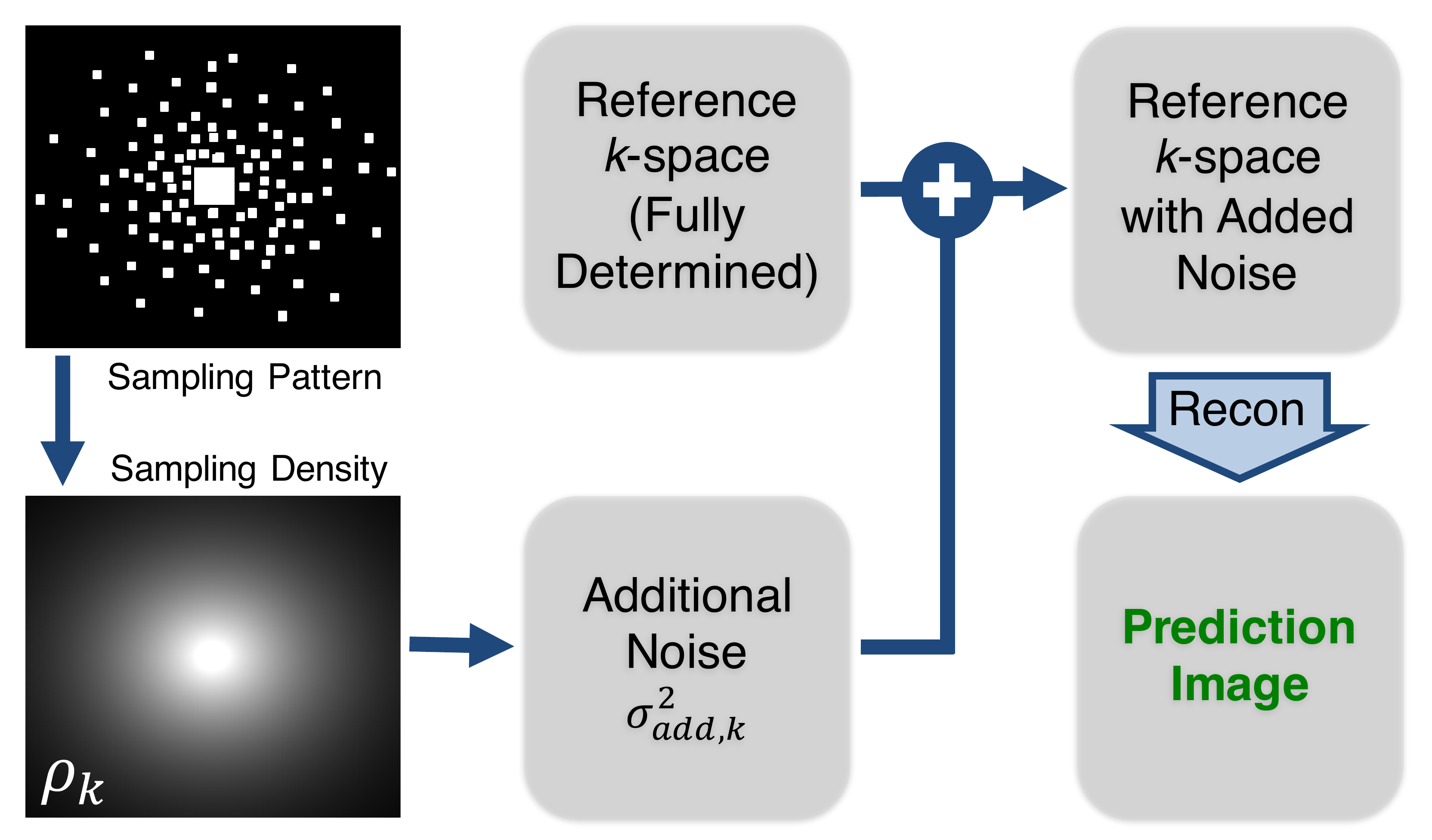}
		\caption{Prediction of Image Quality: The process to add the proper amount of noise to fully-sampled reference $k$-space and reconstruct an image affected by lower SNR due to reduced acquisition time but not affected by an underdetermined system. The expected measurement time at each $k$-space location, $\tau_k$, associated with the given sampling pattern is used to calculate the amount of noise (zero mean, complex Gaussian with variance $\sigma_{add,k}^2$) to add to each position in the reference $k$-space. This $k$-space with added noise is then processed by the reconstruction algorithm to produce the prediction image.}
		\label{fig:prediction}
	\end{center}
\end{figure}

\begin{figure}[!t]
	\begin{center}
		\includegraphics[width=\columnwidth]{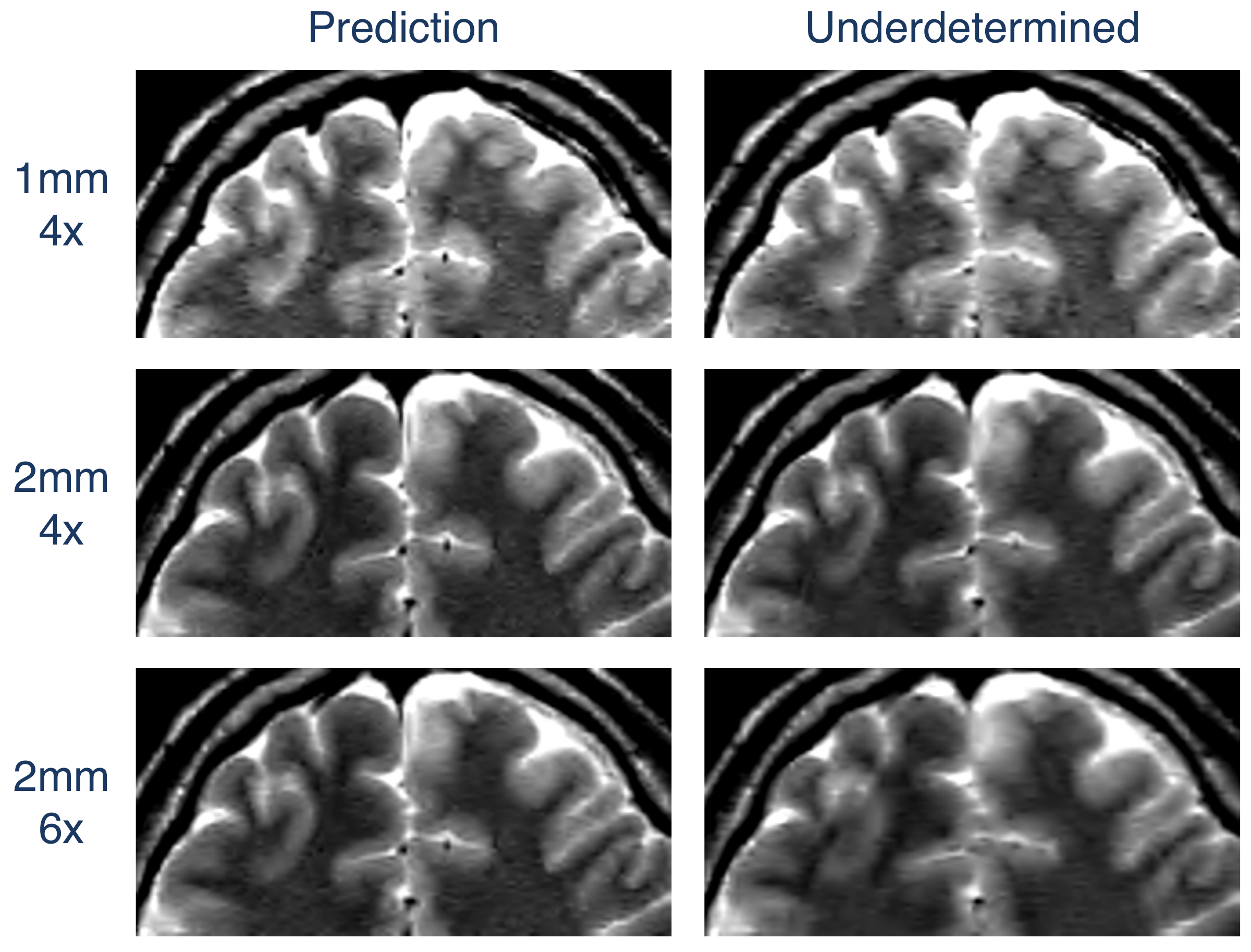}
		\caption{Using the image quality prediction process to adjust scan parameters. This 2D fast spin echo acquisition with 1 mm slice thickness and 4x under-sampling produces poor reconstruction image quality (top right). The corresponding prediction image (top left) also has poor image quality, indicating that noise is the limiting factor. Increasing to 2 mm slice thickness (center row) reduces the noise and produces higher image quality in both the prediction and the underdetermined reconstruction. Further accelerating the scan with 6x under-sampling (bottom row), the prediction image quality is significantly higher than the reconstruction image quality, indicating that the underdetermined system is the limiting factor for those scan parameters.}
		\label{fig:brain_thickness}
	\end{center}
\end{figure}

While it is important to have theory showing that reconstruction techniques are mathematically founded, when attempting to reconstruct a new under-sampled clinical dataset and the image results are unacceptable, it is difficult to leverage the theoretical bounds to understand the cause of the failure. Conversely, when an under-sampled reconstruction is successful at a certain under-sampling rate, it is natural to then ask, how much further can we push under-sampling? In this case, it is difficult to translate theoretic analysis, such as phase transition curves, into practice.
Our goal in this paper is to provide the tools to empirically analyze the effect of lower SNR from reduced measurement time using a reconstruction system that is fully determined, rather than underdetermined. To this end, we present the \textbf{image quality prediction} process (Fig. \ref{fig:prediction}). 
The image quality prediction process takes a Nyquist-sampled (fully determined) reference dataset and adds the proper amount of noise in order to mimic the lower SNR produced by a given under-sampling pattern. By reconstructing this noisy, but still Nyquist-sampled dataset, we have a \textbf{prediction image} that has been affected by lower SNR from reduced measurement time but not by artifacts from an underdetermined reconstruction. The image quality prediction process give us the following three benefits:
\begin{itemize}
\item Comparing the prediction image to the reference reconstruction allows us to see the impact of lower SNR from reduced measurement time on the reconstruction system.
\item Comparing the prediction image to the underdetermined reconstruction, we are able to assess the added effect of the underdetermined system on the reconstructed image.
\item The prediction image provides a better estimate of under-sampled image quality than over-optimistically comparing an underdetermined reconstruction to a fully-sampled reference reconstruction.
\end{itemize}
As exemplified in Fig. \ref{fig:brain_thickness}, for a given clinical application and under-sampling pattern, we can use the image quality prediction process to determine if low SNR, rather than the underdetermined system, is the limiting factor for a successful reconstruction. Specifically, an unsatisfactory prediction image indicates that the under-sampled acquisition contains more noise than the reconstruction can handle. On the other hand, a high quality prediction image and poor results from the underdetermined reconstruction indicate that the constraints on the underdetermined system are not adequate for the limited number of samples acquired.
Once we understand the limiting factor in a given under-sampling application, we can then move forward to modify the acquisition technique to adjust the measurement SNR or the under-sampling rate. We can also focus our efforts on improving the reconstruction algorithm to better account for the noise distribution or to improve the reconstruction constraints, such as the sparsity model.

The remainder of this paper is structured as follows:
Section \ref{sec:theory} lays out the theory related to the image quality prediction process, beginning with the relationship between measurement time and SNR, and ending with a general purpose optimization framework that reconstructs the prediction data as well as under-sampled data.
Sections \ref{sec:methods} and \ref{sec:results} describe our methods and results for our experiments on 1) the effect of measurement time distribution on image quality, 2) the effects of measurement noise and under-sampling rate on image quality, and 3) the image quality prediction process applied to \emph{in vivo} datasets.
Section \ref{sec:discussion} concludes our analysis and records our discussion of limitations and future directions.

\section{Theory} \label{sec:theory}

Before describing the details of the image quality prediction process, we first specify how measurement time affects SNR, specifically when under-sampling. We complete this section by introducing a weighted least squares optimization that generalizes the reconstruction process for both under-sampled data and the fully determined prediction data.

\subsection{Measurement Time and SNR} \label{sec:SNR}

For MRI reconstruction, we can model the signal $s$ as the discrete Fourier transform of the unknown target image object $\boldsymbol{m}$:
\begin{align}
  & s_k = (F\boldsymbol{m})_k 
  \label{eqn:signal_ideal}
\end{align}
where $F$ is the multidimensional discrete Fourier transform operator and $k$ is the $k$-th location in $k$-space. However, each measurement $s_k$ comes with an associated noise. We can model the noisy measurement $Y_k$ as:
\begin{align}
  & Y_k \sim \mathcal{N}\Big(Re(s_k)\text{, }\sigma^2_{acq}/\tau_k\Big) + i\mathcal{N}\Big(Im(s_k)\text{, }\sigma^2_{acq}/\tau_k\Big)
  \label{eqn:signal_measured}
\end{align}
where $Y_k$ is a random variable drawn from a complex-valued Gaussian distribution with mean $s_k$ and variance defined by the system noise variance, $\sigma^2_{acq}$, scaled by one over the measurement time, $\tau_k$, as described in \cite{Macovski}. With this definition, we assume that the signal is deterministic based on our model, the signal is independent from the noise, and that the noise is i.i.d. In cases where these assumptions do not hold, additional care may be taken to adjust the data to this model, for example, pre-whitening coil channels in parallel imaging or accounting for echo time variation in fast spin echo acquisitions.

Again following \cite{Macovski}, we define SNR as the signal intensity divided by the standard deviation of the noise and note that from (\ref{eqn:signal_measured}) we see that the SNR for measured data at the $k$-th location in $k$-space scales with $1/\sqrt{\tau_k}$:

\begin{align}
  SNR = \frac{signal}{\sqrt{noise\text{ }var.}} &= \frac{s_k}{\sqrt{\sigma_{acq}^2/\tau_k}}
  \label{eqn:SNR}
\end{align}
As an example, if we double measurement time at each $k$-space location (e.g. acquire two samples rather than one), the modeled signal remains the same, the noise variance is reduced by a factor of 2, and the SNR increases by a factor of $\sqrt{2}$.

We model the measurement time at the $k$-th location in $k$-space, $\tau_k$, as the acquisition time per sample times the number of samples:
\begin{align}
  & \tau_k = \tau_{acq} n_k
  \label{eqn:tau}
\end{align}
Without loss of generality, we assume a fixed acquisition time for every sample, $\tau_{acq}$, defined by the acquisition parameters and  the number of samples, $n_k$, that may vary across $k$-space locations.

The measurement time $\tau_k$ is not necessarily equal for all $k$-space locations. Variable density sampling across $k$-space can be a natural effect of certain acquisition techniques, such as radial sampling. Variable density sampling may also be desired to take advantage of the higher energy in the low frequency regions to improve SNR (similar to Weiner filtering) or to account for asymptotic incoherence \cite{Adcock}.
A variable density distribution of measurement time generates a corresponding distribution of expected noise variance across $k$-space. Lower sampling density at the high frequency $k$-space locations results in higher variance at these locations, generating a colored (blue) noise distribution, rather than the white noise associated with a uniform acquisition time distribution. It is this colored noise that is coupled with the underdetermined system effects during image reconstruction.

Note that in this paper, we are not attempting to determine the optimal sampling density, but rather provide a tool to help analyze the effects of the chosen sampling density as well as other acquisition and reconstruction parameters.

\subsection{Under-sampling and Expected Measurement Time} \label{sec:US_Noise}

\begin{figure*}[!t]
	\begin{center}
		\includegraphics[width=1.0\textwidth]{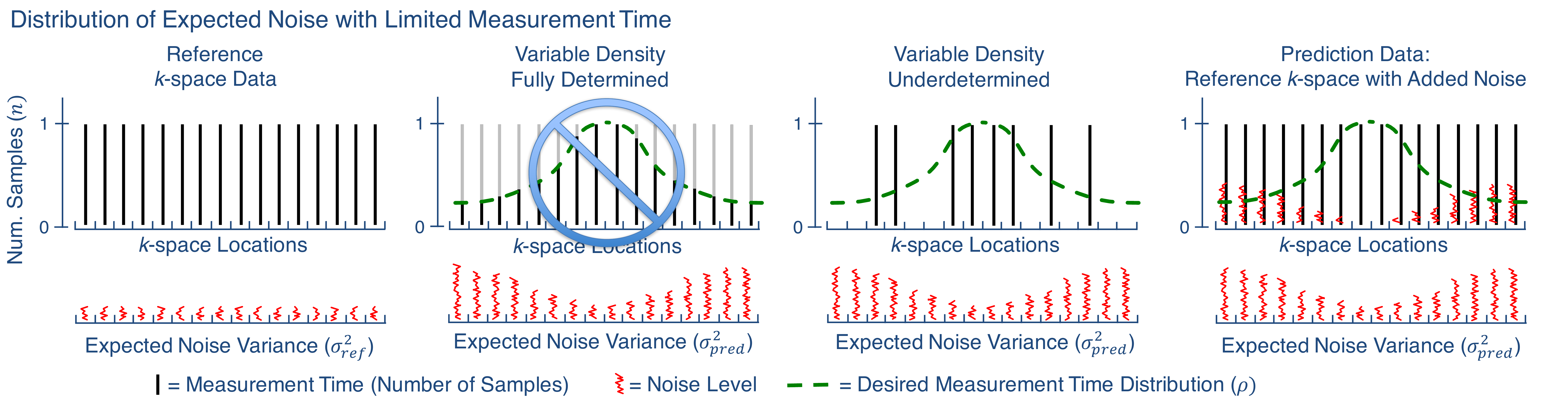}
		\caption{With limited measurement time, the sampling density distribution $\tau$ (dashed green line) may fall below one unit of measurement time. For systems with a minimum measurement time, fractional samples (second column) are not possible, and we are forced to sample below the Nyquist rate (third column) to meet the required measurement time limit. To simulate the infeasible Nyquist-sampled, fully determined acquisition (second column), the image quality prediction process adds noise to a fully determined reference acquisition (fourth column). Note that all three of these datasets have the same distribution of expected noise variance across $k$-space (bottom row).}
		\label{fig:acq_time_dist}
	\end{center}
\end{figure*}

Fast and/or short acquisitions require a limit on the total measurement time. Unfortunately, some systems and applications have constraints on the minimum measurement time at a single $k$-space location. In this case, it is not feasible to sample $k$-space such that the reconstruction system is fully determined (Fig. \ref{fig:acq_time_dist}, second column). Under-sampling is required in order to meet the measurement time constraints without sacrificing other scan requirements, such as spatial resolution, that are defined by the desired measurement time distribution (Fig. \ref{fig:acq_time_dist}, dashed green curve). Without loss of generality, we will define the system's minimum measurement time to be $\tau_k=\tau_{acq}$ and thus any fractional samples, $n_k < 1$, are infeasible.

Under-sampling (Fig. \ref{fig:acq_time_dist}, third column) avoids acquiring fractional samples by measuring either one or zero samples at each $k$-space location. A binary under-sampling pattern can be constructed to fit the desired sampling density, whether it be uniform or variable density. This technique of constructing a continuous output with discrete inputs is analogous to pulse-width modulation in digital signal generation and to digital halftoning in computer graphics.

At first, it may appear that the SNR using these under-sampling patterns is the same as a fully-sampled acquisition because at the $k$-space locations where we collect a measurement, it has the same variance, $\sigma_{acq}^2$, as any fully-sampled measurement. Also, at locations where we don't measure any signal, we also don't collect any noise. However, to understand how the sampling density governs the SNR of the acquired data, we must consider the expected measurement time at each $k$-space location, $\tau_k$.

We model the expected measurement time for random under-sampling patterns by considering the generation of a random sampling pattern. The binary value for each location in the pattern may be determined by drawing a random sample from a Bernoulli distribution. To generate a pattern with a particular sampling density, the mean parameter of each Bernoulli distribution is set to the desired fractional measurement time, $\rho_k$, for that location. Specifically, let us model $T_k$ as a Bernoulli random variable representing the measurement time at a single location in $k$-space. The expected value of $T_k$ is $\tau_{pred,k}$:
\begin{align}
Sample_k &\sim Bern(\rho_k) \label{eqn:sample_1}\\
T_k &= \tau_{acq}n_kSample_k \label{eqn:sample_2}\\
\tau_{pred,k} &= \mathbb{E}[T_k] = \tau_{acq}n_k \rho_k\label{eqn:sample_3}
\end{align} 
$\tau_{pred,k}$ gives us the expected measurement time per $k$-space location, which in turn leads us to the expected noise variance per $k$-space location, $\sigma_{pred,k}^2 = \sigma_{acq}^2/\tau_{pred,k}$.

\subsection{Image Quality Prediction} \label{sec:PIQ}

Using the expected measurement time described in the previous section, the image quality prediction process generates an image that shows the empirical effect of reduced measurement time without any effects of an underdetermined system caused by under-sampling. This process, as depicted in Fig. \ref{fig:prediction}, creates the prediction image by adding noise (based on the expected measurement time of a specific under-sampling pattern) to a fully-sampled reference $k$-space dataset and then passing that adjusted $k$-space through the regularized weighted least squares reconstruction algorithm described in the following section.

The first step in the prediction process is to determine the expected measurement time at each $k$-space location, $\tau_{pred,k}$, for the given under-sampling pattern. For random sampling patterns, this sampling density distribution is readily available, as it is the same distribution that generated the sampling pattern. When the sampling density is not explicitly or analytically available, the measurement time distribution may be approximated from the sampling pattern with local averaging, Voronoi diagrams, or other techniques used in sampling density compensation.

From the measurement time distribution, we calculate how much noise needs to be added to the fully sampled (fully determined) reference $k$-space dataset to match the equivalent statistical noise produced by the given under-sampling pattern. To simulate an under-sampled acquisition with Gaussian noise variance $\sigma_{pred,k}^2=\sigma_{acq}^2/\tau_{pred,k}$, we simply add complex-valued Gaussian noise to the reference $k$-space based on the expected measurement time distribution, $\tau_{pred,k}$ (\ref{eqn:sample_3}), (Fig. \ref{fig:acq_time_dist}, right). Given that $\sigma_{ref}^2=\sigma_{acq}^2/\tau_{ref}$ is the Gaussian noise variance measured from the reference data, we can calculate the variance of the complex Gaussian noise, $\sigma_{add,k}^2$, to add to location $k$ in the reference $k$-space:
\begin{align}
\sigma_{pred,k}^2 &= \sigma_{ref}^2 + \sigma_{add,k}^2 \label{eqn:add_3} \\
\sigma_{add,k}^2 &= \sigma_{pred,k}^2 - \sigma_{ref}^2 \label{eqn:add_4} \\
  &= \bigg(\frac{\tau_{ref}}{\tau_{pred,k}} - 1 \bigg) \sigma_{ref}^2 \label{eqn:add_5}
\end{align}
where $\tau_{ref} = \tau_{acq}n_{ref}$ (\ref{eqn:tau}) and $n_{ref}$ is the number of samples acquired in the reference data. The detailed derivation between (\ref{eqn:add_4}) and (\ref{eqn:add_5}) may be found in the appendix, section \ref{sec:appendix_sigma_add}. Often $n_{ref}=1$, however, the reference data may be acquired using many samples, for example the number of averages might equal two or, in the case of our first two experiments, $n_{ref}=144$ (Fig. \ref{fig:acq_time_setup}).

Note that with variable density sampling patterns, $\tau_{pred,k}$, is not constant across $k$-space, and thus, the variance of the added noise, $\sigma_{add,k}^2$, will also vary across $k$-space.

The noise to add at each point in $k$-space is drawn from a complex-valued, zero-mean Gaussian distribution with variance equal to the $\sigma_{add,k}^2$ for that $k$-space location. This noise is simply added to the reference $k$-space to produce fully determined $k$-space with the noise distribution matching that of the under-sampled data (see Fig. \ref{fig:acq_time_dist}, right).

The final step in the image quality prediction process is to pass the reference $k$-space with added noise through the regularized weighted least squares reconstruction algorithm described in the following section, producing a prediction image that gives an estimate of the reconstruction image quality assuming no effect from an underdetermined reconstruction system.

\subsection{Weighted Least Squares Reconstruction} \label{sec:WLS}

We require a consistent reconstruction formulation that supports standard fully-sampled and under-sampled data as well as the prediction data. To this end, we use a maximum \textit{a posteriori} (MAP) formulation of MRI reconstruction that leads, in general, to a regularized weighted least squares optimization. Equations (\ref{eqn:signal_ideal}) and (\ref{eqn:signal_measured}) combine to give us a Gaussian likelihood probability of measuring a signal $y_k$ given an image object $\boldsymbol{m}$:
\begin{align}
P(y_k|\boldsymbol{m}) = \frac{1}{\sqrt{2\pi\sigma_{acq}^2/\tau_k}} exp\bigg(\frac{-|y_k - (F\boldsymbol{m})_k |^2}{2\sigma_{acq}^2/\tau_k}\bigg)
\end{align}
With this Gaussian likelihood and assuming a general prior probability on our image data $P(\boldsymbol{m})$, the resulting MAP formulation leads to a weighted-least squares optimization:
\begin{align}
\boldsymbol{m}^* &= \argmax_{\boldsymbol{m}} P(\boldsymbol{m}|\boldsymbol{y}) \label{eqn:WLS_1} \\
  &= \argmax_{\boldsymbol{m}} P(\boldsymbol{y}|\boldsymbol{m}) P(\boldsymbol{m}) \label{eqn:WLS_2} \\
  &= \argmin_{\boldsymbol{m}} \frac{1}{2} \sum_{k=1}^{N_P} \tau_k |y_k-(F\boldsymbol{m})_k|^2 - \log P(\boldsymbol{m}) \label{eqn:WLS_3} \\
	&= \argmin_{\boldsymbol{m}} \frac{1}{2} ||W\boldsymbol{y}-WF\boldsymbol{m}||_2^2 - \log P(\boldsymbol{m}) \label{eqn:WLS_4}
\end{align}
where $\boldsymbol{m}$ is the vectorized image with $N_P$ number of pixels; $\boldsymbol{y}$ is the vectorized acquired $k$-space locations with $N_P$ number of elements; $F$ is the $N_P$x$N_P$ multidimensional discrete Fourier transform operator; and $W$ is an $N_P$x$N_P$ diagonal matrix with $W_{k,k} = \sqrt{\tau_k}$ values along the diagonal. The detailed derivation between (\ref{eqn:WLS_2}) and (\ref{eqn:WLS_3}) may be found in the appendix, section \ref{sec:appendix_WLS}.

In this paper, we will use a Laplacian-based prior to promote sparsity, ($-\log P(\boldsymbol{m}) = \lambda||\Psi(\boldsymbol{m})||_1$), where $\Psi$ is a sparsity transform function and $\lambda$ is the Laplace prior parameter. This $\ell_1$ regularized WLS optimization does not have an analytic solution, and finding the solution requires a non-linear reconstruction algorithm. In general, we can solve this optimization using an iterative algorithm, such as FISTA \cite{FISTA} or ADMM \cite{ADMM}.

This optimization framework, given the proper weight values described below, generalizes the reconstruction of a) fully sampled, b) under-sampled, and c) image quality prediction datasets.

\begin{enumerate}[a)]
\item \textbf{Fully sampled weights:}
The least squares weights for a fully sampled dataset (both uniform and variable density sampling) are simply equal to the square root of the measurement time, $W_{k,k} = \sqrt{\tau_k} = \sqrt{\tau_{acq} n_k}$, for the $k$-th sample. Assuming, again, that the acquisition time per sample is constant across $k$-space, $\tau_{acq}$ may be pulled out of the $\ell_2$ norm term, simplifying the weights to be equal to the square root of the number of samples, $W_{k,k} = \sqrt{n_k}$.

Note that with $n_k$ constant across $k$-space and a uniform prior $P(\boldsymbol{m})$, the MAP optimization becomes the standard least squares optimization:
\begin{align}
\boldsymbol{m}^* = \argmin_{\boldsymbol{m}} \frac{1}{2} ||\boldsymbol{y}-F\boldsymbol{m}||_2^2
\end{align}

\item \textbf{Under-sampled weights:}
When under-sampling, the weights, $W_{k,k}$, are simply set to one or zero depending on whether or not that $k$-space location has been sampled (assuming the same measurement time at each sampled location). With these binary weights, the operator $W$ in (\ref{eqn:WLS_4}) becomes the under-sampling operator defined by the binary sampling pattern. With a Laplacian-based prior, the MAP reconstruction becomes the standard Lasso optimization \cite{Lasso} commonly used in compressed sensing. In addition to strictly binary under-sampling patterns, the WLS optimization also allows for under-sampling patterns that have zero measurement time at certain locations and a range of measurement times across the remaining locations, for example, an acquisition with under-sampled high frequencies and over-sampled low frequencies.

\item \textbf{Prediction weights:}
The prediction data is designed to simulate the noise variance from the expected measurement time for a given sampling density $\rho_k$, leading us to WLS weights $W_{k,k}= \sqrt{\tau_{pred,k}} = \sqrt{\tau_{acq} n_{ref,k} \rho_k}$, which may be simplified to $W_{k,k} = \sqrt{\rho_k}$ assuming constant sampling time and constant number of samples per location in the fully-sampled reference data.

\end{enumerate}

\section{Methods} \label{sec:methods}

\subsection{Effect of Measurement Time Distribution}

\begin{figure*}[!t]
	\begin{center}
		\includegraphics[width=\fullfigwidth\textwidth]{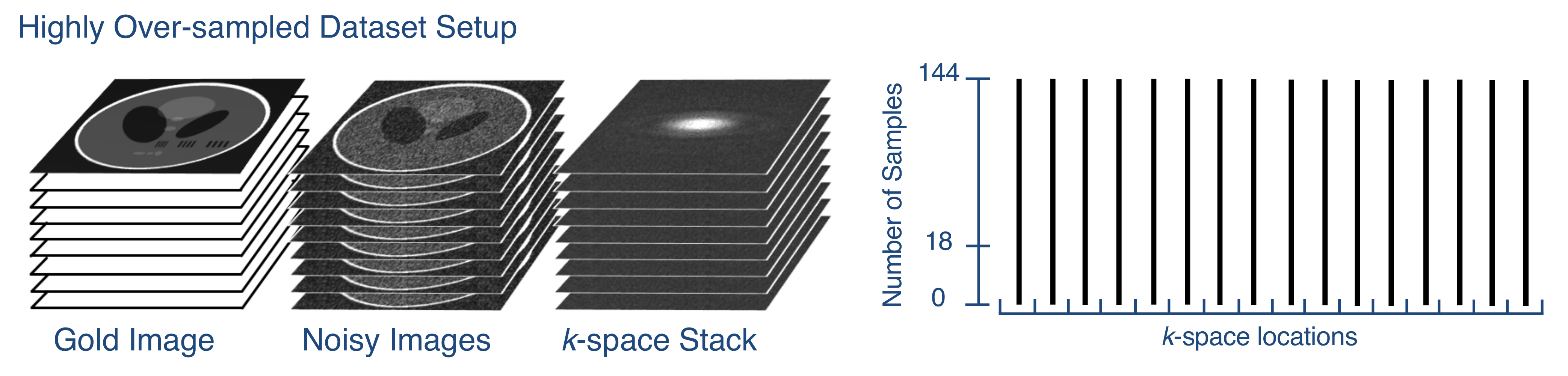}
		\caption{Experimental setup allowing us to choose the number of acquisition samples (from 0 to 144) at each $k$-space location. Noise is added to 144 copies of the input gold image. These noisy images are then Fourier transformed to created a stack of $k$-space images with 144 samples available at each $k$-space location. Note that for the tomato dataset, there is no gold image and the $k$-space stack comes directly from the 144 scanner acquisitions of the tomato.}
		\label{fig:acq_time_setup}
	\end{center}
\end{figure*}

To better understand effects of reduced measurement time and under-sampling and to test our image quality prediction process, we created an experiment that enables us to compare the reconstructions of 1) a fully determined dataset, 2) an underdetermined dataset, and 3) the corresponding prediction data, all using the same total measurement time and sampling density distribution.

The foundation of this experiment is a "stack" of 144 fully-sampled $k$-space images (Fig. \ref{fig:acq_time_setup}). Each entry in the $k$-space stack is a different noisy acquisition of the same object slice. With 144 samples available at each of the $N$ $k$-space locations, we are able to select a subset of these samples to simulate acquiring a specific number of samples at each $k$-space location based on a desired measurement time distribution.

We used two different datasets for this experiment. The first dataset was the classic Shepp-Logan digital phantom \cite{Shepp_Logan} with a slight modification to add a set of parallel dark bars that will help analyze spatial resolution. This phantom was chosen because it has an explicitly sparse representation (many true zero values) in the finite differences domain (often seen in total variation (TV) reconstructions), implying that we can use compressed sensing find a proper solution to the underdetermined system of equations caused by under-sampling. As seen in Candes, Romberg, and Tao \cite{Candes_Tao}, the Shepp-Logan phantom, without noise, may be perfectly recovered after severe under-sampling. To analyze how noise propagates through the reconstruction system, we generated a different instance of complex-valued, zero-mean, Gaussian noise to add to 144 copies of the $k$-space for the Shepp-Logan phantom. The second dataset is 144 actual MRI acquisitions of a tomato at a single slice location. This data was acquired on a Seimens 3T scanner with a 2D T2-weighted fast spin echo sequence. Only the body coil was used during acquisition to both simplify the reconstruction model and ensure that each of the 144 acquisitions had relatively low SNR.

For both datasets, we selected a subset of the full stack of $k$-space samples based on three different sampling distributions, as depicted in the top row of Fig. \ref{fig:acq_time_results_vardens}: \textbf{reference}, using all $144N$ samples (where $N$ is the number of $k$-space locations); \textbf{fully determined}, selecting only $18N$ samples according to either a uniform or variable density sampling distribution across $k$-space locations; and \textbf{underdetermined}, selecting $18N$ samples and following the same density distribution but collecting all 144 samples at $N/8$ randomly chosen $k$-space locations and collecting zero samples for the remaining locations. We also reconstructed both datasets using the \textbf{image quality prediction} process to add noise to the $144N$ reference dataset to simulate the noise level from the $18N$ fully determined dataset.

For all reconstructions, the selected $k$-space samples were averaged at each $k$-space location to create a single $k$-space image to be reconstructed ($\boldsymbol{y}$, from equations (\ref{eqn:WLS_1}) and (\ref{eqn:WLS_4})).

We reconstructed all data using our implementation of ADMM, formulated for the regularized weighted least squares optimization, with the weights equal to the number of measurements acquired at each $k$-space location, as specified in section \ref{sec:WLS}. For the digital phantom dataset, we used isotropic total variation as the sparsity model. For the single channel MRI acquired data, we used Daubechies-4 wavelets with translation invariant cycle spinning \cite{Cycle_Spinning} as the sparsity model.

\subsection{Effects of Measurement Noise and Under-sampling Rate} 

Given enough acquisition time, we can satisfy a given sampling density distribution by either Nyquist sampling $k$-space or by under-sampling. Both of these sampling patterns produce similar distributions of expected noise variance in our data, but under-sampling incurs an additional cost from having an underdetermined system of equations.
In this experiment, we will extend the over-sampled stack experiment above to take a closer look at the effect of measurement noise and under-sampling rate on reconstruction image quality.
We accomplish this by varying both the measurement noise level and the under-sampling rate and then comparing the MSE images reconstructed from variable density \textbf{fully determined} data and from variable density \textbf{underdetermined} data.

As in the measurement time experiment above, we have a stack of $k$-space data, and we generate an output image by reconstructing a subset of $k$-space samples, selected according to a either a fully determined variable density sampling pattern or an underdetermined pattern following the same measurement time distribution. 

In this experiment, the $k$-space stack is generated from copies of a single relatively high SNR (31.3 dB) \emph{in vivo} head acquisition. Similar to the Shepp-Logon $k$-space stack, we added to  $k$-space a sample of complex-valued, Gaussian noise with zero mean and a given standard deviation. We executed the experiment using three different values for the added noise standard deviation (1, 5, 8) and four under-sampling rates (2x, 4x, 8x, and 12x under-sampled). 

The head dataset for this experiment is an axial slice of a 3D fully-sampled, spoiled gradient echo dataset acquired on a 1.5T GE Healthcare scanner with 8 receive channels. This multi-channel dataset was preprocessed, using ESPIRiT coil sensitivity maps \cite{ESPIRIT}, to combine the data into a single-channel, allowing us to use a simpler reconstruction model for this experiment. This head dataset has relatively high SNR, so we are able to experiment with very low noise and with increased noise as we add various amounts of noise the $k$-space stack for this dataset. This head dataset also provides a real example of an image that is only approximately sparse in the wavelet transform domain.

We repeated these 12 experiments (three noise levels by four under-sampling rates) 100 times, each time reconstructing the fully determined data and the underdetermined data, as well as the corresponding prediction data. We then plotted the resulting MSE values (relative to the original head image) (Fig. \ref{fig:acq_time_grid}).

\subsection{Image Quality Prediction}

We demonstrate the image quality prediction process by comparing the output of the actual under-sampled reconstruction to both the generated prediction image and the fully determined reference image. We executed this experiment for two \emph{in vivo} fully-sampled MRI datasets using increasingly aggressive retrospective under-sampling rates.

\subsubsection{\emph{in vivo} Knee}

The \emph{in vivo} knee dataset is an axial slice of a 3D fully-sampled, fast spin echo dataset acquired on a 3T GE Healthcare scanner with 8 receive channels. This dataset was collected by Epperson et al \cite{Knee_Repository} and is available at \cite{MRI_Data_org}.

The two retrospective under-sampling patterns used were 4x and 12x under-sampled, variable density Poisson disc. Both patterns fully-sampled the center of $k$-space to allow for ESPIRiT auto-calibration \cite{ESPIRIT}. Neither the reference data nor the under-sampling patterns included the corners of $k$-space, a common acquisition acceleration.

The optimization equation for this parallel imaging, compressed sensing reconstruction is an extension of equation (\ref{eqn:WLS_4}), modified to include parallel imaging and a Laplacian prior:

\begin{align}
  & \min_{\boldsymbol{m}} \frac{1}{2}\|W\boldsymbol{y} - WFS\boldsymbol{m}\|^2_2 + \lambda \|\Psi \boldsymbol{m}\|_1
  \label{eqn:opt_brain}
\end{align}
where $\boldsymbol{m}$ is the vectorized image with $N_P$ number of pixels; $\boldsymbol{y}$ is the vectorized acquired multi-channel $k$-space data with $N_CN_P$ number of elements ($N_P$ is the number of pixels, $N_C$ is the number of coils); $F$ is the $N_CN_P$x$N_CN_P$ 2D Fourier transform operator for each coil independently; $S$ is the $N_CN_P$x$N_P$ block diagonal sensitivity maps generated with ESPIRiT calibration; $\Psi$ is the sparsity transform; $\lambda$ is the regularization parameter; and $W$ is the $N_CN_P$x$N_CN_P$ diagonal weight matrix.

Note that the reconstruction process now includes the parallel imaging coil combination operator $S^H$. With the addition of parallel imaging, the under-sampled reconstruction system is now both ill-conditioned and underdetermined. The image quality prediction process in this case will empirically show the effect of lower SNR due to reduced measurement time on the compressed sensing and parallel imaging reconstruction without any effect from an underdetermined or ill-conditioned system. The actual underdetermined reconstruction will then produce an image affected by similar lower SNR as well as the effects from the ill-conditioned and underdetermined parallel imaging and compressed sensing system.

The sparsity filter (associated with $\Psi$) used within the reconstruction was wavelet soft-thresholding using Daubechies-4 with translation invariant cycle spinning \cite{Cycle_Spinning}.

The regularized weighted least squares optimization for both prediction and underdetermined reconstruction used our implementation of the ADMM algorithm. The only difference between the two reconstructions was the appropriate change to the weights as specified in section \ref{sec:WLS}. Specifically, the weights for the prediction reconstruction were the square root of the sampling density, $\rho_k$, at each $k$-space location, and the actual underdetermined reconstruction weights were binary with ones for acquired locations and zeros elsewhere.

The image quality prediction process requires an understanding of the existing noise level in the fully-sampled reference data ($\sigma_{full}^2$). Ideally, this noise level could be obtained from a explicit measurement of the received signal using the coils on the same scanner, prior to the actual exam. In our experiments, we measured the noise level from the reference data directly by Fourier transforming the (multi-channel) $k$-space data and measuring the variance of the values from a 11x11 background patch in each coil image. The noise level was measured and applied independently for each coil channel.

A direct inverse 2D Fourier transform followed by coil combination ($\boldsymbol{m}_{ref} = S^HF^{-1}\boldsymbol{y}_{ref}$) was used on the reference $k$-space to generate the fully-sampled reference image for comparison (Fig. \ref{fig:knees} top).

\subsubsection{\emph{in vivo} Head}

The \emph{in vivo} head dataset is an axial 2D fast spin echo dataset acquired on a 3T Siemens Trio scanner with 12 receive channels. The 12 coil channels were reduced to 4 channels with Siemens coil compression. Multiple slices were acquired at slice thicknesses of 1 mm and 2 mm. The phase encode lines were retrospectively undersampled at 4x and 6x acceleration using a 1D variable density poisson disc sampling with the center 24 lines fully sampled. This dataset was processed in the same manner as the \emph{in vivo} knee dataset above.

\section{Results} \label{sec:results}
The following three results are shared across all of our experiments:

\begin{enumerate}
\item the \textbf{prediction} image has equivalent or worse image quality than the \textbf{reference} image,
\item the \textbf{under-sampled} reconstruction image has equivalent or worse image quality than the \textbf{prediction} image,
\item the \textbf{prediction} image for a given sampling density has equivalent image quality to the \textbf{fully determined} image with the same sampling density.
\end{enumerate}

\subsection{Effect of Measurement Time Distribution}

\begin{figure*}[!t]
	\begin{center}
		\includegraphics[width=\fullfigwidth\textwidth]{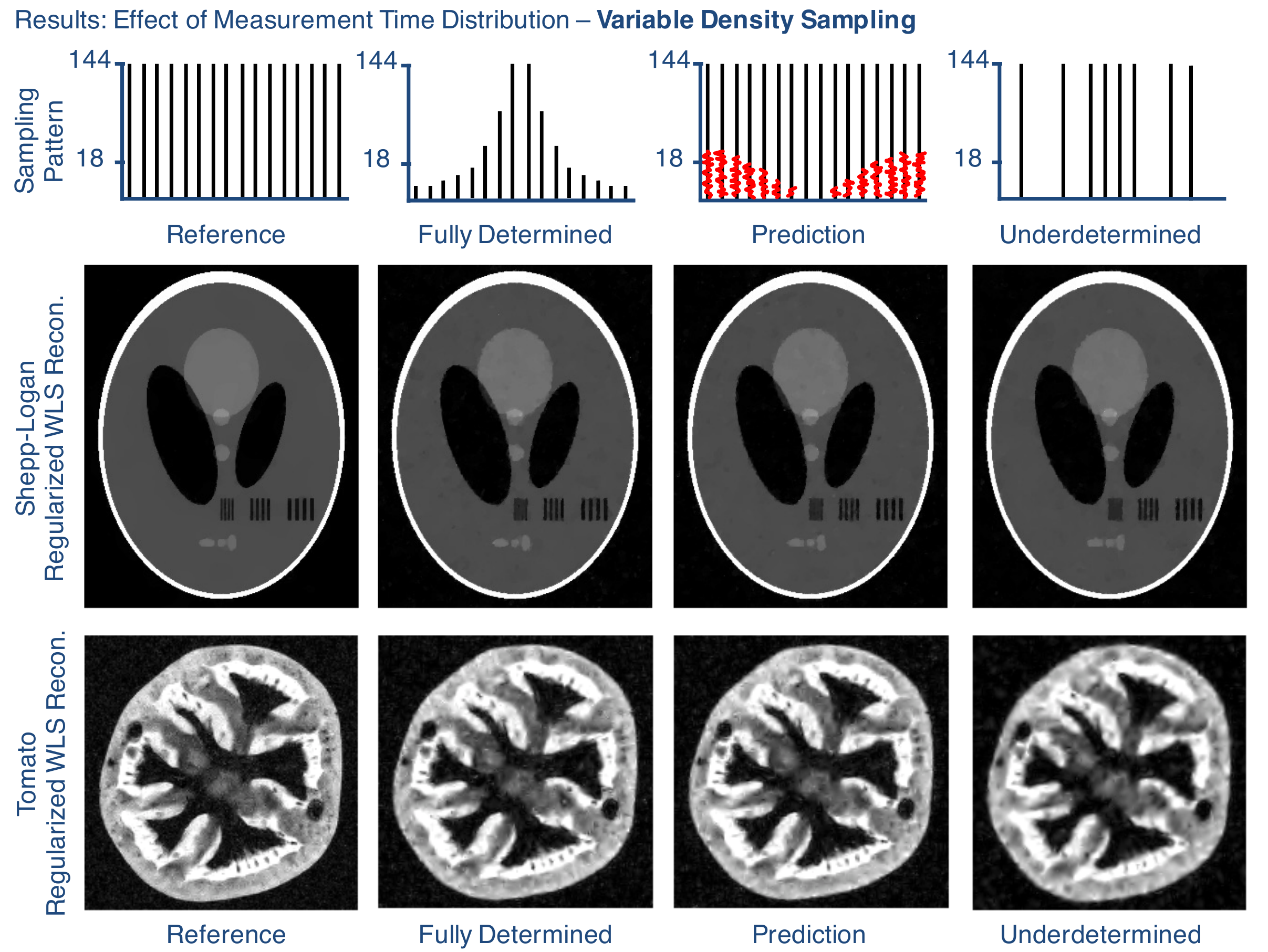}
		\caption{Results from the effect of the measurement time distribution experiment using variable density sampling. Each column uses a different set of $k$-space samples; from left to right: \textbf{over-sampled reference}, using all 144 samples at each $k$-space location; \textbf{variable density, fully determined}, using 1/8 of the total samples following a variable density distribution; \textbf{prediction data}, using fully-sampled $k$-space with noise added to simulate the variable density, fully determined dataset; \textbf{variable density (randomly sampled), underdetermined}, using 1/8 of the total samples and following the same variable density distribution, but only using either 144 or zero samples at each location. \emph{Row 1}: Illustration of how measurement time is distributed across $k$-space. \emph{Row 2}: WLS reconstruction of the Shepp-Logan data, regularized with total variation. \emph{Row 3}: WLS reconstruction of the tomato $k$-space data, regularized with wavelets.}
		\label{fig:acq_time_results_vardens}
	\end{center}
\end{figure*}

Fig. \ref{fig:acq_time_results_vardens} shows the results of our experiment to test the effect of various measurement time distributions on reconstruction image quality. For both the Shepp Logan digital phantom and the MRI acquisition of the tomato, the fully determined images with reduced measurement time show lower image quality than the images reconstructed from the reference acquisition data. As seen specifically in the blurred spatial vertical bars, the fully determined images did not fully recover from the limited acquisition time despite not having any corruption from an underdetermined systems of equations.

The variable density underdetermined Shepp Logan reconstruction (Fig. \ref{fig:acq_time_results_vardens}, third row, right) was successful and has nearly identically image quality to the fully determined reconstruction but still lower image quality than the reference reconstruction. This indicates that the underdetermined reconstruction recovered well from the underdetermined system, but still could not completely recover from the lower SNR due to reduced measurement time. For the acquired tomato dataset, however, the underdetermined image quality (Fig. \ref{fig:acq_time_results_vardens}, bottom, right) is lower than the prediction and fully determined image quality, indicating that the reconstruction could not completely recover from the underdetermined system. This is not a surprising result because the tomato image is not perfectly sparse in the wavelet transform domain, especially when compared to the explicit sparsity of the Shepp Logan phantom in the finite differences domain.

Fig. \ref{fig:acq_time_results_vardens} also shows the results of the image quality prediction process for the same two datasets and sampling distributions. The second and third columns in this figure show that the fully determined reconstructions have essentially identical image quality to their corresponding prediction images. This verifies that the image quality prediction process closely simulates the noise level and reconstructed image quality of the associated fully determined acquisitions.


\subsection{Effects of Measurement Noise and Under-sampling Rate} \label{sec:Results_NoiseRate}

\begin{figure}[!t]
	\begin{center}
		\includegraphics[width=\colfigwidth\columnwidth]{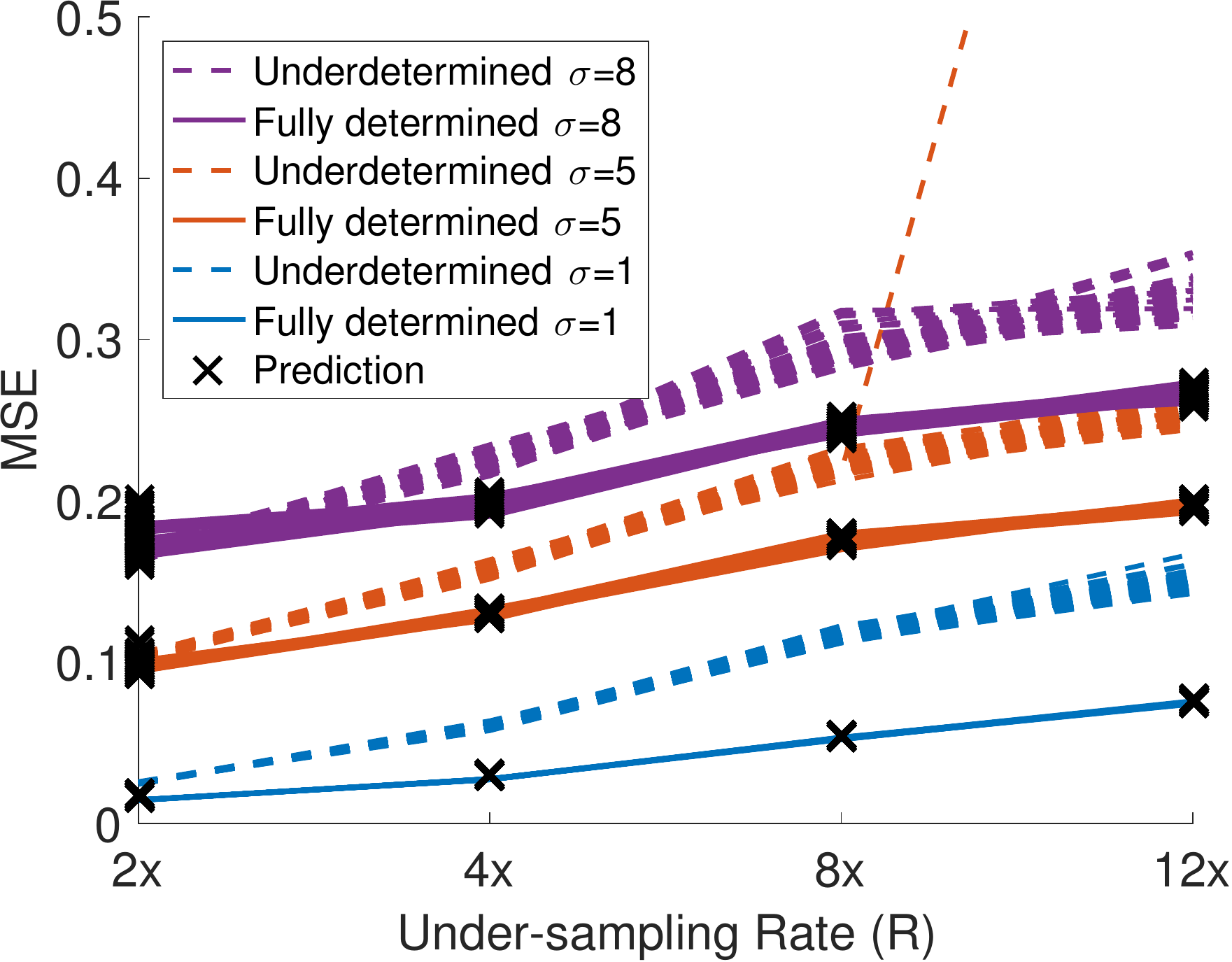}
		\caption{Results of our experiment to compare fully determined and underdetermined reconstructions with the same total measurement time across four different under-sampling rates (2x, 4x, 8x, 12x) for three different noise levels (added noise standard deviations 1.0, 5.0, 8.0). Mean squared error (MSE) values are plotted for each of the 100 repetitions of the same experiment. Note that for the 2x under-sampling rate, the fully determined and underdetermined reconstructions have essentially the same MSE. As the under-sampling rate increases, the underdetermined system produces an increasingly worse MSE than the fully determined system. Note that one of the 100 underdetermined reconstructions at $\sigma=5$ and $R=12x$ failed to converge. This outlier is consistent with compressed sensing theory and practice where the reconstruction may fail to converge at higher under-sampling rates.}
		\label{fig:acq_time_grid}
	\end{center}
\end{figure}

By varying the input noise level and the under-sampling rate, we see the differences in the resulting MSE for the reconstructions of the fully determined data and underdetermined data, both with the same measurement time distribution. Fig. \ref{fig:acq_time_grid} shows that for a fixed noise level and increasing under-sampling rate, the MSE of the fully determined images increases, showing that the reduced measurement time affects image quality despite no under-sampling. Also, as we increase the under-sampling rate, the MSE of the underdetermined images increases significantly faster than the fully determined images. This gap in image quality shows the degrading effect of the underdetermined reconstruction increasing as the under-sampling rate increases and the sparsity transform can no longer adequately model the image in a sufficiently sparse representation.

As seen in Fig. \ref{fig:acq_time_grid}, the MSE of the prediction images matches the MSE of the fully determined reconstructions for all noise levels and under-sampling rates, indicating that the image quality prediction process is consistently simulating the expected noise level for the given sampling density.

The results from this experiment help us to see that when the image quality of the prediction image is unacceptable, the actual under-sampled reconstruction will also be unacceptable (i.e. higher MSE). In this situation, the low SNR of the acquisition is the limiting factor in the reconstruction, not the artifacts due to the underdetermined system. To improve the reconstruction in this case, steps should be taken adjust the acquisition parameters to increase the SNR or better handle the expected noise levels (e.g. reducing spatial resolution, decreasing under-sampling rate, or improving the image prior $P(\boldsymbol{m})$).

\subsection{Image Quality Prediction}

\begin{figure}[!t]
	\begin{center}
		\includegraphics[width=0.8\columnwidth]{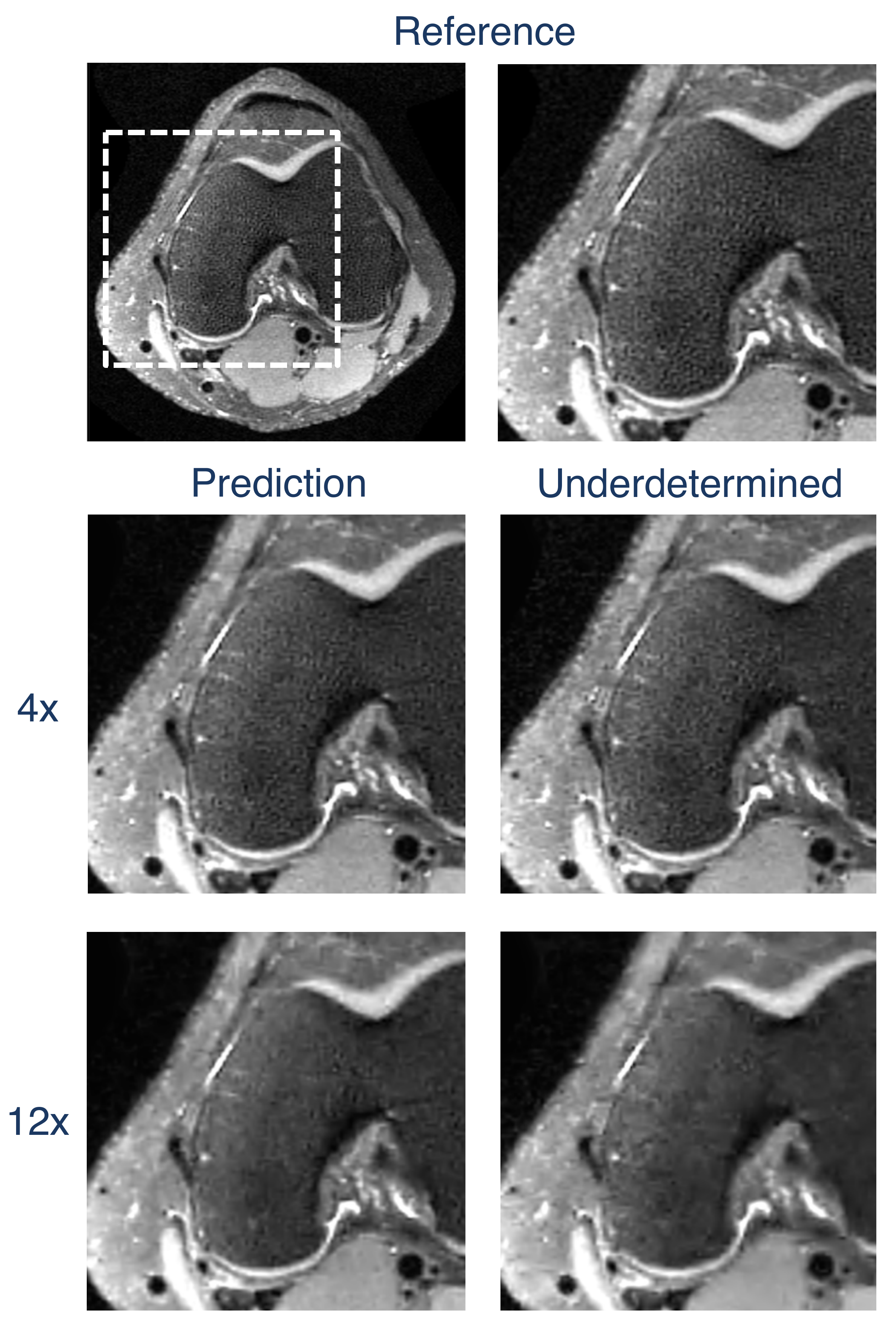}
		\caption{Results from \emph{in vivo} image quality prediction experiment. Fully determined reference axial knee (top) followed by prediction (left column) and actual underdetermined reconstruction (right column) for two different under-sampling rates: 4x and 12x. Images are zoomed and cropped to show image quality detail.}
		\label{fig:knees}
	\end{center}
\end{figure}

Fig. \ref{fig:brain_thickness} shows the prediction and underdetermined reconstruction images for the \emph{in vivo} head experiment. This figure illustrates how the prediction image may be used to gain insight into the causes of poor under-sampled image quality and adjust scan parameters, such as slice thickness, as needed.

Fig. \ref{fig:knees} shows the reference, prediction, and underdetermined reconstruction images for the \emph{in vivo} experiment using the knee dataset and various under-sampling rates. Fig. \ref{fig:knees} shows the following three qualitative results: 1) the reference image has better image quality than the prediction images; 2) the prediction images have better image quality than the corresponding underdetermined images; and 3) the underdetermined images are more similar in image quality to the prediction images than the reference image. That the reference images look better than the prediction images is expected because the prediction process adds more noise to the fully sampled reference data. That the prediction images look better than the underdetermined image is expected because the underdetermined reconstruction had to find a proper solution to an underdetermined system of equations in addition to recovering from the lower SNR from reduced measurement time. Finally, the prediction image provides a better estimate of reconstruction image quality than the reference image.

With a reasonable amount of under-sampling, the 4x underdetermined images only have slightly lower image quality than the prediction images. As under-sampling increases to 12x, the image quality gap between the underdetermined and prediction images increases. These results are consistent with our effect of measurement noise and under-sampling rate experiment when increasing sampling rate (section \ref{sec:Results_NoiseRate}).

\section{Conclusion} \label{sec:discussion}

The presented \textbf{image quality prediction} process provides an empirical upper bound on under-sampled image quality, which serves as better metric for evaluating the effectiveness of a reconstruction algorithm than direct comparison to a fully-sampled reference reconstruction. The prediction process enables an analysis of a reconstruction algorithm's ability to handle lower SNR due to reduced measurement time without any effect from an underdetermined system. By simulating the effect of lower SNR without any underdetermined effects, the prediction process allows us to determine whether a reconstruction is actually limited by our sparse recovery or simply limited by low acquisition signal to noise ratio. Comparison of the prediction image to the reference reconstruction provides a means to assess the effect of lower SNR on reconstruction image quality. Comparison of the prediction image to the underdetermined reconstruction enables us to analyze what artifacts are introduced when under-sampling is used rather than fully determined following the same measurement time distribution. The image quality prediction results and analysis are consistent with our experiments using our highly over-sampled datasets to explicitly compare reconstruction results from variable density fully determined and underdetermined data.

An additional benefit of the prediction process is that it may be used to compare and tune different reconstruction algorithms or parameters, assessing how different reconstruction handle the lower SNR due to reduced measurement time in addition to comparing the actual under-sampled reconstructions.

A limitation of the image quality prediction process is that it requires a fully-sampled reference dataset. Access to a fully-sampled acquisition is not always possible, especially in cases with 3D or 4D dynamic imaging, where long, fully-sampled acquisition times are not practical. The image quality prediction process is also limited in the fact that it can isolate the effect of low SNR without the effects from an underdetermined system, but it does not isolate these effects without the lower SNR due to reduced measurement time. Future work could investigate the effects of an underdetermined systems using \emph{in vivo} data by reconstructing fully-sampled reference datasets that are highly over-sampled to have minimal input noise. Of course, the effects of the underdetermined system would still be dependent on the image content, which varies significantly across clinical applications. 

While developing the image quality prediction process, we use a maximum \textit{a posteriori} formulation to derive a \textbf{general weighted least squares optimization framework} that accounts for both uniform and variable density sampling patterns, with under-sampling as a special case using binary weights. This weighted least squares formulation adjusts the standard least squares term to account for the colored noise arising from the distribution of expected measurement time across $k$-space locations. Future work could develop methods to similarly incorporate the effect of measurement time distribution into the sparsity regularization term, allowing the sparsity filters to better recover from colored noise in addition to incoherent aliasing. 

\section{Appendix}

\subsection{Variance of Added Gaussian Noise} \label{sec:appendix_sigma_add}

Derivation between equations (\ref{eqn:add_4}) and (\ref{eqn:add_5}) to determine the variance of the Gaussian noise to add during the image quality prediction process:

\begin{align}
\sigma_{add,k}^2 &= \sigma_{under,k}^2 - \sigma_{ref}^2 \tag{\ref{eqn:add_4}} \\
  &= \frac{\sigma_{acq}^2}{\tau_{pred,k}} - \sigma_{ref}^2 \\
  &= \frac{\tau_{ref}}{\tau_{ref}}\frac{\sigma_{acq}^2}{\tau_{pred,k}} - \sigma_{ref}^2 \\
  &= \frac{\tau_{ref}}{\tau_{pred,k}}\sigma_{ref}^2 - \sigma_{ref}^2 \\
  &= \bigg(\frac{\tau_{ref}}{\tau_{pred,k}} - 1 \bigg) \sigma_{ref}^2 \tag{\ref{eqn:add_5}}
\end{align}

\subsection{Weighted Least Squares} \label{sec:appendix_WLS}

Derivation of the maximum likelihood formulation of MRI reconstruction optimization. This derivation applies directly to the likelihood term of the MAP derivation described in section \ref{sec:WLS}, specifically between equations (\ref{eqn:WLS_2}) and (\ref{eqn:WLS_3}):

\begin{align}
& \boldsymbol{m}_{MLE}^* \\
  &= \argmax_{\boldsymbol{m}} P(\boldsymbol{y}|\boldsymbol{m}) \\
  &= \argmin_{\boldsymbol{m}} -\log P(\boldsymbol{y}|\boldsymbol{m}) \\
  &= \argmin_{\boldsymbol{m}} -\log \prod_{k=1}^{N_P} P(y_k | \boldsymbol{m}) \\
  &= \argmin_{\boldsymbol{m}} -\sum_{k=1}^{N_P} \log  P(y_k | \boldsymbol{m}) \\
  &= \argmin_{\boldsymbol{m}} -\sum_{k=1}^{N_P} \log  \frac{1}{\sqrt{2\pi\sigma_{acq}^2/\tau_k}} exp\bigg(\frac{-|y_k - s_k|^2}{2\sigma_{acq}^2/\tau_k}\bigg) \\  
  &= \argmin_{\boldsymbol{m}} \sum_{k=1}^{N_P} \log  \frac{-1}{\sqrt{2\pi\sigma_{acq}^2/\tau_k}} +\frac{|y_k - (F\boldsymbol{m})_k|^2}{2\sigma_{acq}^2/\tau_k} \\  
  &= \argmin_{\boldsymbol{m}} \frac{1}{2}\sum_{k=1}^{N_P} \tau_k |y_k - (F\boldsymbol{m})_k|^2 \\  
\end{align}

\section*{Acknowledgment}

This research was made possible with support from the United States Department of Defense National Defense Science \& Engineering Graduate Fellowship (NDSEG), National Institutes of Health (NIH) R01EB009690 grant, GE Healthcare, and a Sloan Research Fellowship. Thank you to Frank Ong and Jonathan Tamir for great discussions about noise, optimization algorithms, and reconstruction software.

\bibliographystyle{IEEEtran}
\bibliography{IEEEabrv,mybib}

\end{document}